\documentclass[10pt,twocolumn,letterpaper]{article}

\usepackage{cvpr}
\usepackage{times}
\usepackage{epsfig}
\usepackage{graphicx}
\usepackage{amsmath}
\usepackage{amssymb}
\usepackage{bm}
\usepackage{subfigure}

\usepackage[pagebackref=true,breaklinks=true,letterpaper=true,colorlinks,bookmarks=false]{hyperref}

\cvprfinalcopy 

\def\cvprPaperID{1170} 

\ifcvprfinal\fi
\begin{document}
	
	\title{Facelet-Bank for Fast Portrait Manipulation}
	\author{Ying-Cong Chen$^1$ \quad Huaijia Lin$^1$ \quad Michelle Shu$^3$ \quad Ruiyu Li$^1$ \quad Xin Tao$^1$ \\
		Xiaoyong Shen$^2$ \quad Yangang Ye$^2$ \quad Jiaya Jia$^{1,2}$ \\
		$^1$The Chinese University of Hong Kong \quad $^2$Tencent Youtu Lab \quad $^3$Johns Hopkins University \\
		{\tt\small \{ycchen, linhj, ryli, xtao\}@cse.cuhk.edu.hk}
		\quad
		{\tt\small goodshenxy@gmail.com}
		\\ 
		{\tt\small Mshu1@jhu.edu}
		\quad
		{\tt\small yangangye@tecent.com}
		\quad
		{\tt\small leojia9@gmail.com}
	}

	\maketitle
	
	\begin{abstract}
		Digital face manipulation has become a popular and fascinating way to touch images with the prevalence of smart phones and social networks. With a wide variety of user preferences, facial expressions, and accessories, a general and flexible model is necessary to accommodate different types of facial editing. In this paper, we propose a model to achieve this goal based on an end-to-end convolutional neural network that supports fast inference, edit-effect control, and quick partial-model update. In addition, this model learns from unpaired image sets with different attributes. Experimental results show that our framework can handle a wide range of expressions, accessories, and makeup effects. It produces high-resolution and high-quality results in fast speed.
		
	\end{abstract}
	
	\section{Introduction}
	
	Digital face manipulation aims to change semantically expressive and meaningful attributes, such as smiling and mourning, or add virtual makeup/accessories to human faces, including mustache and eyeglasses. With the increasing popularity of smart phones and digital cameras, the demand of a practical and fast system rises drastically.
	Face manipulation has attracted great interests in computer vision and graphics \cite{kemelmacher2014illumination, blanz2003reanimating, cao2014facewarehouse, blanz1999morphable, yang2011expression, wang2009face, thies2015real}. Previous methods devoted to face beautification \cite{leyvand2006digital,chen2009automatic}, de-beautification \cite{chen2017makeup}, expression manipulation \cite{thies2015real} and age progression \cite{kemelmacher2014illumination}, to name a few.
	
	Given these many solutions, it is common knowledge that different face makeup or attribute changes require special manipulation operations. For example, face beautification or de-beautification processes skin color and texture while face expression manipulation focuses more on 2D or 3D geometry. With this fact, most approaches were particularly designed for individual tasks where any specialization requires expert efforts and domain knowledge to establish new solutions for effect generation.
	
	In the following, we elaborate on the underlying problems of face manipulation when seeking a data-driven framework to unify many face effects. It is followed by introducing the intriguing work to construct our system with this pursuit.
	
	\begin{figure}
		\centering
		\includegraphics[width=1\linewidth]{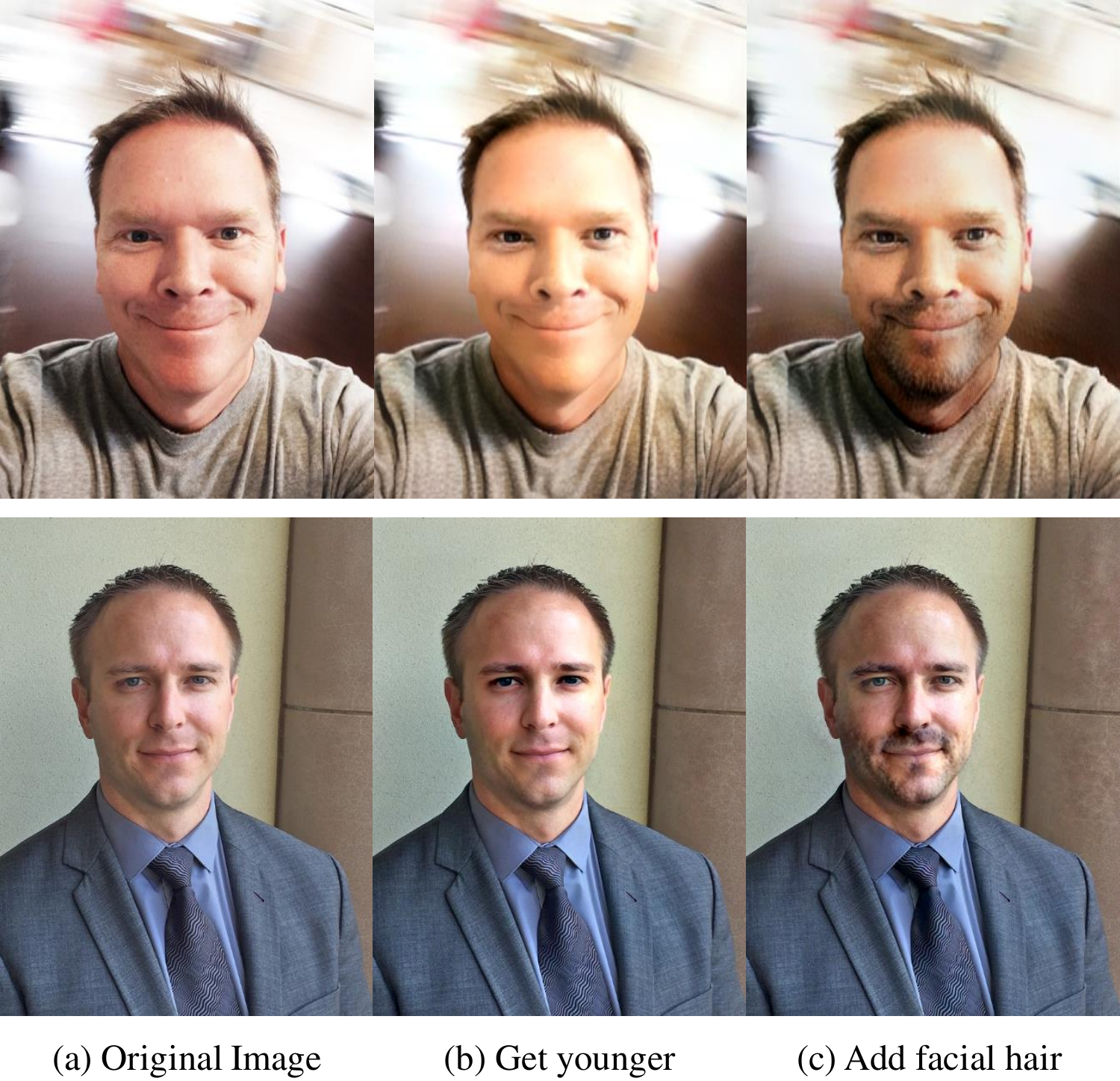}
		\caption{Illustration of face manipulation using our model. } \label{fig:teaser}
	\end{figure}
	
	\subsection{Possible Solutions and Their Problems}
	
	\paragraph{Direct Regression}
	The straightforward way to learn face editing operations from external data is to directly regress the input (before edit) and the ground truth image after edit \cite{chen2017makeup,brock2016neural}. However, this procedure requires labeled paired data, which in many cases is unavailable or requires intensive human labor to create. For any effects that do not exist before, these operations cannot be established easily.
	
	\paragraph{Generative Adversarial Network}
	Recently, Generative Adversarial Network (GAN) has shown its capacity in set-to-set unsupervised learning \cite{CycleGAN2017}. It uses a cycle consistent loss to preserve image content, and an adversarial loss to transfer the attribute of one set to another. Although the concept is neat and results are quite phenomenal, it is difficult to train, especially for new effects that require system component modification. The training needs to keep a balance of generation and discrimination. We notice non-optimal training would result in unrealistic manipulation, which could be easily noticeable on visually-sensitive human faces.
	
	\paragraph{Deep Feature Interpolation}
	Deep feature interpolation \cite{upchurch2016deep}  provides another solution to learn image attribute change from two different sets. It is based on the deep feature of two image sets.
	However, it is not an end-to-end framework and thus cannot be optimized globally. In addition, it is computation-intensive even during testing because of its heavy involvement of hundreds of face warping and convolution operations.
	
	\subsection{Our Solution}
	
	We pursue a general, flexible and high-quality-output network for face manipulation. Fig. \ref{fig:teaser} shows the results generated by our method. Our work follows the an encoder-decoder architecture rather than the popular GAN. Inspired by the Style-Bank \cite{chen2017stylebank} that learns replaceable style transfer layers, we propose a Facelet-Bank framework that models face effects with respective middle-level convolutional layers. So interestingly, in order to generate different effects, instead of redesigning the framework completely, only the middle-level convolutional layers need to be updated.
	
	Also, considering lack of ground truth data for many face manipulation tasks, we leverage the result of \cite{upchurch2016deep} to produce pseudo targets to learn the facelet-bank layers. The local receptive field of our facelet bank naturally provides regularization, and thus it can capture the correct relation between visual patterns and certain network operations, despite the fact that the pseudo targets are usually noisy.
	
	Finally, we show that these layers can automatically attend to the most important regions so that face manipulation can be performed in an end-to-end fashion. Our method is specially designed to allow users to control the level of effect, and thus it enables interactive face manipulation. Our overall contribution is multifold.
	
	\begin{itemize}
		\item We propose a set-to-set CNN framework for face manipulation. It does not requires paired data in training.
		\item The framework is flexible to generate different effects and their levels by simply updating a few convolutional layers, which makes the system developer-friendly.
		\item Our method naturally benefits from the local prior of convolutional layers, which regularizes noisy labels.
		\item Experiments show that our approach can handle a wide range of face effects in fast speed.
	\end{itemize}

	\begin{figure*}
		\centering
		\includegraphics[width=1\linewidth]{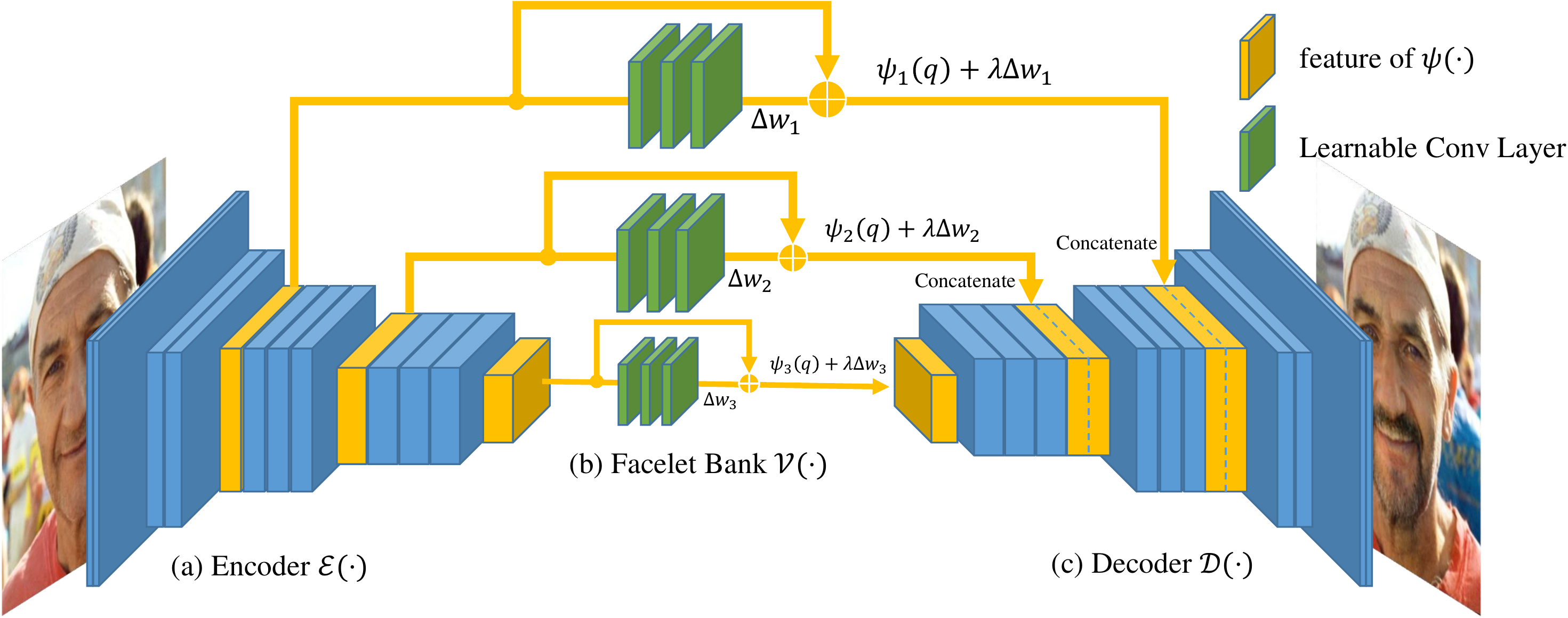} 
		\caption{Illustration of our framework. (a) is the encoder $\mathcal{E}(\cdot)$; (b) are convolutional layers of our facelet bank $\mathcal{V}(\cdot)$; (c) is the decoder $\mathcal{D}(\cdot)$. The structure of our facelet bank is Conv-ReLU-Conv-ReLU-Conv, where all Convs are with $3\times 3$ kernels. Also, all Convs of the facelet bank do not change the height, width and number of channels given the previous input.}
		\label{fig:framework}
	\end{figure*}
	
	\section{Related Work}
	
	\paragraph{Face Editing} 
	Our work can be categorized as face editing or manipulation, which has been extensively studied in computer vision and graphics \cite{wang2009face, blanz2003reanimating, yang2011expression, blanz1999morphable, cao2014facewarehouse, kemelmacher2014illumination}. Traditional face edit includes 
	face relighting \cite{wang2009face,blanz2003reanimating},
	expression editing \cite{yang2011expression},
	face morphing \cite{blanz1999morphable},
	attribute editing \cite{cao2014facewarehouse},
	and face aging \cite{kemelmacher2014illumination}. 
	However, these models are designed for specific tasks, and rely heavily on domain knowledge. Ours method differs from them as it is a data-driven framework designed to handle general face effects and manipulation.

	\paragraph{Image Attribute Manipulation} 
	Altering semantic attributes of images is an important topic. Image analogy \cite{hertzmann2001image, cao2012registration, wang2006deringing} transfers the attribute or appearance changes from a pair of images to a new one. 
	Recently, generative adversarial networks \cite{isola2016image, CycleGAN2017, NeuralFace2017, domain_transfer,perarnau2016invertible} have become popular in image generation and image-to-image transform. 
	Specifically, the method in \cite{isola2016image} learns attribute transformation from two sets of paired data. This idea is further generalized to handle unpaired data \cite{CycleGAN2017,domain_transfer,NeuralFace2017}.
	These approaches alter the attributes by training a generator to simulate an online updating domain discriminator. However, when designing and training a system, the balance between generator and discriminator is difficult to deal with. Scaling up image sizes would make training even more difficult. Besides, these approaches do not allow control over the degree of editing, e.g., how bright the face is, which is very important to real-world applications.

	\paragraph{Feature Interpolation} 
	Our work is also related to feature interpolation  \cite{larsen2015autoencoding, radford2015unsupervised, upchurch2016deep,yan2016attribute2image,kingma2013auto,yeh2016semantic}.
	Variational autoencoder (VAE) \cite{larsen2015autoencoding, yeh2016semantic} learns a latent space such that image manipulation can be done with simple arithmetic computation in the deep space. But it implicitly assumes that the target attribute is well isolated. Given unpaired data for face manipulation, this assumption is usually not satisfied.
	The method of \cite{upchurch2016deep} alleviates this problem by proposing a K-NN based approach to isolate the target attribute. Yet only a subset of samples are used for estimation, which lead to undesired visual artifacts in resulted images. In addition, the computational cost is inevitably high, which may be intolerable for fast face manipulation.
	Our work is different from these approaches. We use convolutional neural networks to learn the shifting direction, so that each transformation is associated with a corresponding visual pattern. We show that the resulting direction is better isolated towards the target attribute. In addition, our method is an end-to-end framework that supports fast inference.
	
	\section{Proposed Method}
	
	Suppose there are two face domains $\mathcal{X}$ and $\mathcal{Y}$ with different properties. Our goal is to shift images of domain $\mathcal{X}$ towards domain $\mathcal{Y}$ without guidance from  any paired data.
	As for face manipulations, we want the algorithm to produce intermediate results so that users can control the strength of operations, such as smoothness of the skin.
	
	This process can be represented as  
	\begin{equation} \label{eq:shift:1}
	z = \mathcal{O}(x, \lambda),
	\end{equation}
	where $x$ and $z$ are the input and output images, $\lambda$ controls the strength, and $\mathcal{O}$ denotes the operation that changes $x$ from domain $\mathcal{X}$ toward domain $\mathcal{Y}$.
	
	Note that domains $\mathcal{X}$ and $\mathcal{Y}$ usually differ \textit{semantically}, so $\mathcal{O}$ can be highly complex. In order to simplify $\mathcal{O}$, we instead define the operation $\mathcal{O}'$ in a deep space, and change Eq. \eqref{eq:shift:1} to
	\begin{equation} \label{eq:shift:2}
	\psi(z) = \mathcal{O}'(\psi(x), \lambda),
	\end{equation}	
	where $\psi(\cdot)$ denotes the deep space. To one extreme, if $\psi(\cdot)$ captures enough rich semantic information, $\mathcal{O}'$ can be further simplified as linear shift in the deep space \cite{li2017universal}, which leads to
	\begin{equation} \label{eq:shift:2}
	\psi(z) = \psi(x) + \lambda \Delta v,
	\end{equation}		
	where $\Delta v$ denotes the direction from domain $\mathcal{X}$ to domain $\mathcal{Y}$ in the deep space. 
	
	As indicated in \cite{zeiler2014visualizing}, a network trained on large-scale data captures semantic attributes in the convolutional layers.
	This idea is validated in the work of \cite{li2017universal,upchurch2016deep}, which uses a VGG network \cite{vgg} to encode the semantic information for deep feature interpolation or style transfer. We follow this idea to use the $5$ convolutional layers to construct $\psi(\cdot)$. To represent face attributes at different levels,  ReLU3\_1, ReLU4\_1 and ReLU5\_1 are jointly used to represent $\psi(\cdot)$.
	
	\paragraph{Reversing $\psi(\cdot)$} 
	Following \cite{li2017universal,chen2017stylebank}, we train a fixed decoder network $\mathcal{D}$ to decode $\psi(z)$ to obtain the final output. The decoder $\mathcal{D}$ has reverse architecture of the encoder $\mathcal{E}$, except for concatenating the manipulated features to the corresponding layers.
	We use the following loss function to train $\mathcal{D}$:
	\begin{equation} \label{eq:feat_consistent} \small
	\mathcal{L} =   \sum_{i=1}^n ||z_i-x_i||_2^2 + \omega \sum_{i=1}^{n}||\mathcal{E}(x_i) - \mathcal{E}(\mathcal{D}(\mathcal{E}(x_i)))||^2_2,
	\end{equation}
	where the first and second terms impose consistencies in pixel and feature space respectively, and $\omega$ is the weight to balance the two losses. We also observe that pre-training of the decoder with $\mathcal{L}=\sum_{i=1}^{n}||(\mathcal{E}(x_i)+ \lambda \Delta v_i) - \mathcal{E}(\mathcal{D}(\mathcal{E}(x_i)+\lambda \Delta v_i))||^2_2$ is helpful, where $\Delta v_i$ are pseudo labels of different attributes computed with Eq \eqref{eq:delta_v_2}.
	
	\paragraph{Overall Structure}
	The overall network architecture is illustrated in Fig. \ref{fig:framework}, which is composed of an encoder $\mathcal{E}$ that transforms images to a deep space, a ConvNet $\mathcal{V}$ that estimates the domain direction shift $\Delta v$, and a decoder that transforms the operated deep feature back to an image. With this effective pipeline, the encoder and decoder are responsible for general operations, while $\mathcal{V}$ is the key component that determines the specific effect required for a face manipulation. 
	
	Now, realizing differences across all types of face effects, from a smile on the face to adding mustache, no longer needs frequent redesign of the framework. By using different $\mathcal{V}$, these apparently dissimilar face operations can be accomplished accordingly. Thus, we name the collection of $\mathcal{V}$ as \textbf{Facelet Bank}, representing the collection of different face effects we can achieve.
	
	\begin{figure*}
		\centering
		\includegraphics[width=1\linewidth]{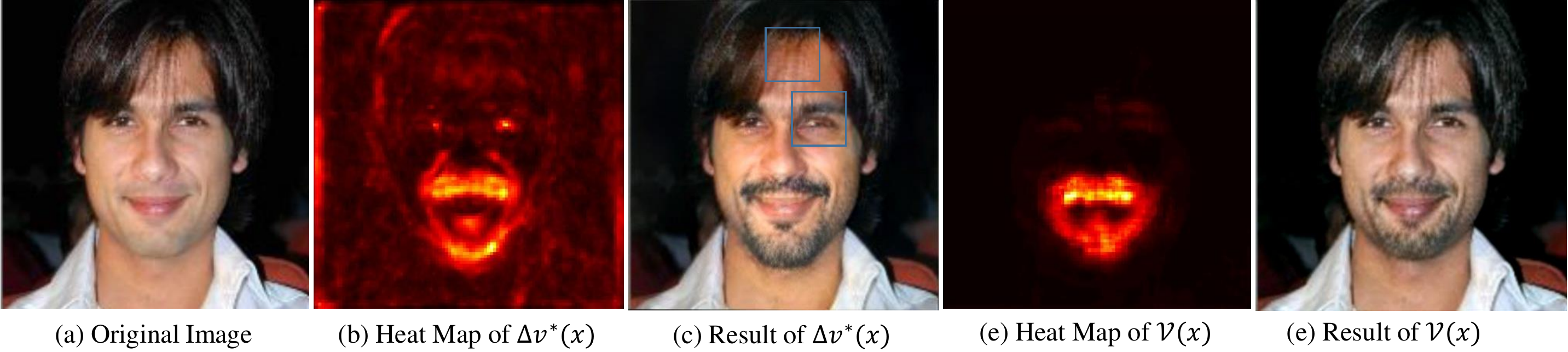}
		\caption{Illustration of the noise resistance effect. (a) The original image. (b) The heat map of pseudo shifting direction computed by Eq. \eqref{eq:shift:2}. The blue rectangles mark the undesired change regions. (c) Corresponding result of Eq. \eqref{eq:shift:2}. (d) Heat map of our estimated direction shift. (e) Our result.}
		\label{fig:denoise}
	\end{figure*}
	
	\subsection{Learning Facelet Bank}
	
	Estimating $\Delta v$ requires cancellation of all factors except for the target attributes.
	Without paired data, a straightforward approach is to compute the average difference between $\mathcal{X}$ and $\mathcal{Y}$ \cite{larsen2015autoencoding, yeh2016semantic}, namely 
	\begin{equation} \label{eq:delta_v_1}
	\Delta v \leftarrow \frac{1}{m} \sum_{i=1}^m \psi(y_i) -  \frac{1}{n} \sum_{i=1}^n \psi(x_i), 
	\end{equation}
	where $m$ and $n$ are the training samples of domains $\mathcal{X}$ and $\mathcal{Y}$ respectively. 
	However, this implies a strong assumption that all training samples have similar attributes to the queried one. When this assumption is not satisfied, the model does not produce realistic results.
	
	\paragraph{Generating Pseudo Labels}
	We relax this assumption by adopting a query-adaptive attribute transform model  $\mathcal{V}(x)$, where $x$ is a query sample.
	
	Note that learning $\mathcal{V}(\cdot)$ is difficult as we do not have paired-data to infer $\Delta v$ for each training sample. Inspired by the method in \cite{upchurch2016deep}, we average neighbors of each training sample of \textit{both} domains to construct the attribute vector. Specifically, for sample $x$, the corresponding $\Delta v$ is computed as
	\begin{equation} \label{eq:delta_v_2}
	\Delta v^{*} (x) = \frac{1}{K} \sum_{i\in N_{\mathcal{Y}}^K(x)} \psi(y_i) -  \frac{1}{K} \sum_{i\in N_{\mathcal{X}}^K(x)} \psi(x_i),
	\end{equation}   
	where  $N_y^K(x)$ and $N_x^K(x)$ refer to $K$ nearest neighbors of $x$ among sets $\mathcal{Y}$ and $\mathcal{X}$ respectively. To reduce the influence of pose, viewpoint and rotation, all training face images are aligned to a frontal face template in advance\footnote{A face alignment tool \cite{kazemi2014one} in DLIB \cite{dlib} is used in our experiment.}. Here the ``average" operation suppresses undesired noise and thus $\Delta v^*$ tends to cancel out other factors except those that divide domains $\mathcal{X}$ and $\mathcal{Y}$. 
	
	\paragraph{Network Design}
	Since $\Delta v^*$ relates strongly to the deep feature, we reuse the extracted deep feature by setting the input of $\mathcal{V}(\cdot)$ to $\mathcal{V}(\mathcal{E}(q))$. We stack 3 fully convolutional layers with ReLU activation to capture the non-linearity relation. In this way, $\mathcal{V}$ can be learned as
	\begin{equation} \label{eq:V}
	\mathcal{L}_{\mathcal{V}} = \sum_{i=1}^n|| \mathcal{V}(\mathcal{E}(x_i)) - \Delta v^{*} (x_i) ||_2^2,  
	\end{equation}
	where $x_i$ refers to training samples of set $\mathcal{X}$. $\Delta v^{*} (x_i) $ is defined in Eq. \eqref{eq:delta_v_2}.

	\subsection{More Analysis of the Facelet Bank} 
	Although our facelet bank learns with the help of pseudo paired data $\Delta v^*$ defined in Eq. \eqref{eq:V}, it has additional advantages compared with directly using $\Delta v^*$.
	
	\paragraph{Noise Resistance} 
	In practice, our facelet bank can even outperform the pseudo label $\Delta v^*$. To demonstrate it, we compute the heat map of $\Delta v$ as 
	\begin{equation} \label{eq:heatmap}
	H_{i,j} = \sum_{k=1}^{c} v_{i,j,k}^2,
	\end{equation}
	where $i=\{1,2,\cdots, h\}$, $j=\{1,2,\cdots,w\}$, and $k=\{1,2,\cdots,c\}$ denotes the row, column and channel indexes respectively. Fig. \ref{fig:denoise} shows the heat map of ``add beard'' operation. Intuitively, adding beard is only relevant to the mouth region, while other place should not be updated. However, the pseudo target $\Delta v^*$ activates in many regions as shown in the heat map. This may cause undesirable changes in wrong places as shown in Fig \ref{fig:denoise}(c). Note that the pseudo label is computed by only a subset of samples\footnote{$2K=200$ samples are used in our method, which is less than $1\%$ compared with the training set.}, not sufficient to cancel out all noise. On the contrary, our facelet bank has a much cleaner heat map, which only activates in the correct region.
	
    Although this may look counterintuitive at the first glance, it actually makes sense if we consider the built-in regularization of the convolutional neural network. Note that CNN has local receptive fields, which force the system to capture the relation between certain visual patterns (the mouth in this case) and the corresponding manipulation operations (``add beard'' in this case). By training on a large set of samples, unrelated activations can be well suppressed, and only the relevant regions are activated.

	\begin{figure}
		\centering
		\includegraphics[width=1\linewidth]{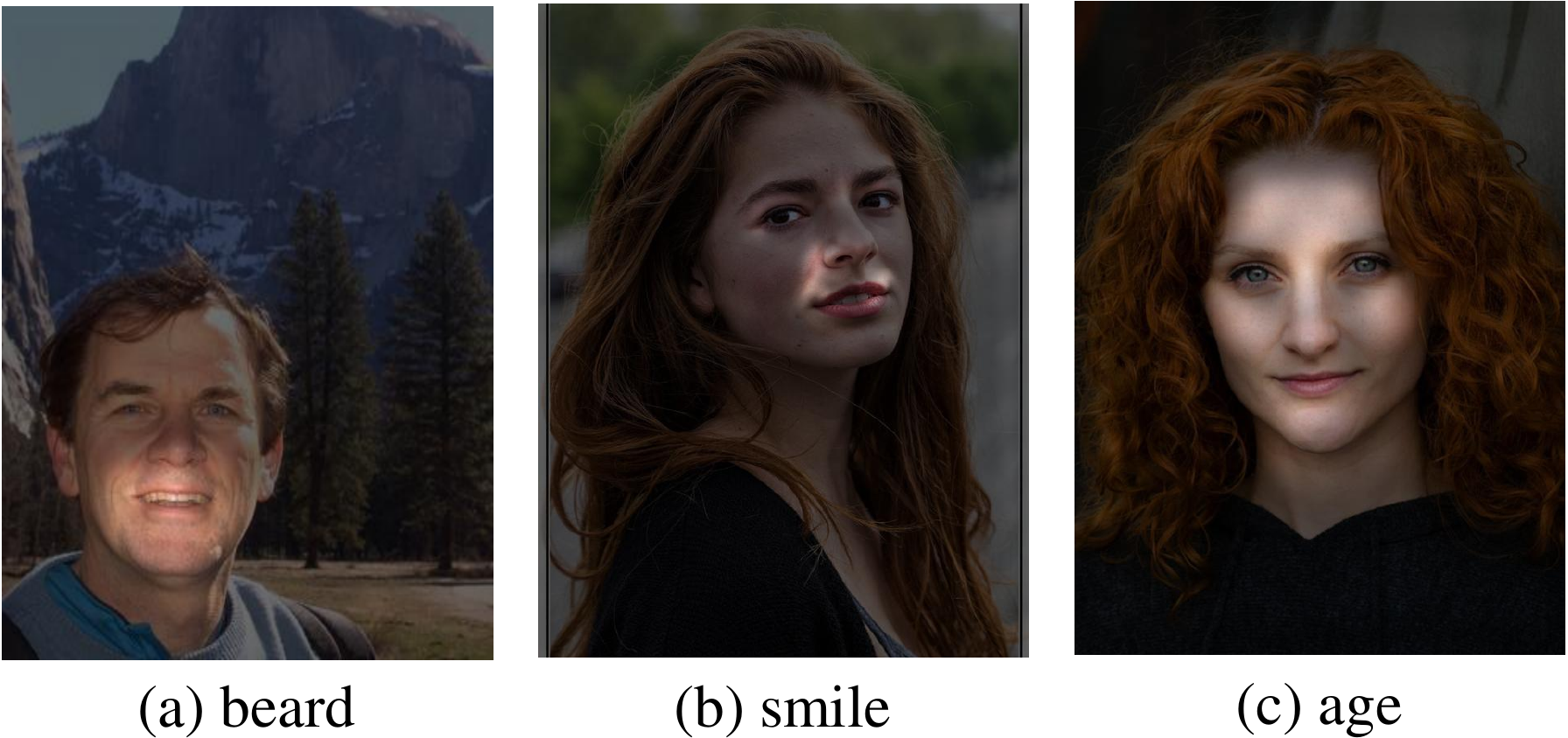} 
		\caption{Visualization of the attention region. The attention mask is computed by Eq. \eqref{eq:heatmap}. (a), (b) and (c) correspond to the operations of adding beards, making smiles and changing age. Note that for the beard effect, our facelet bank focuses on the mouth area. For the smile effect, it attends to smile-related facial muscles. As for the age changing effect, the attention region covers the whole face. These results match our intuition.} \label{fig:attention}
	\end{figure}

    \begin{figure*}[h!]
	\centering
	\subfigure[ Original Image ]{\includegraphics[width=0.24\textwidth]{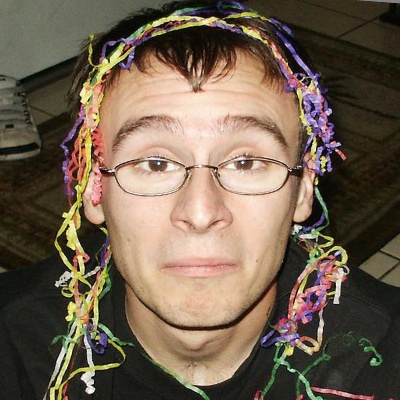}}
	\subfigure[ Global Mean ]{\includegraphics[width=0.24\textwidth]{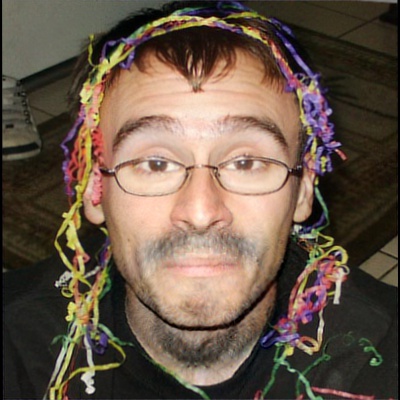}}	    
	\subfigure[ K-NN Mean ]{\includegraphics[width=0.24\textwidth]{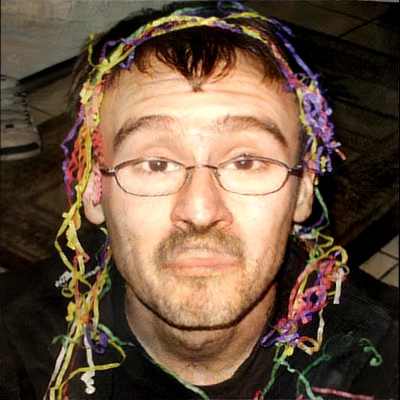}}	   
	\subfigure[ Ours  ]{\includegraphics[width=0.24\textwidth]{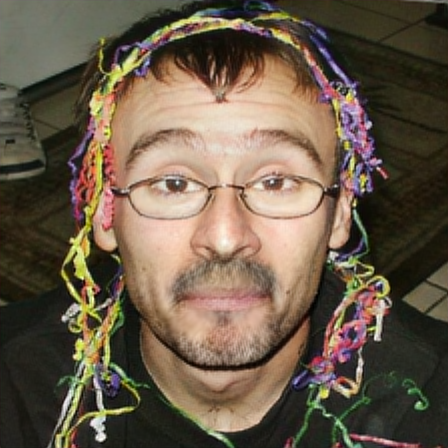}}
	\caption{Comparing our facelet-bank method with baseline approaches. {Please zoom in to see details.}}
	\label{fig:exp1}
    \vspace{0.1in}
    	\centering
    	\subfigure[ Original Image ]{\includegraphics[width=0.24\textwidth]{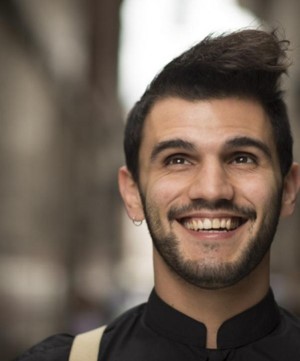}}
    	\subfigure[ Layer 5 Only  ]{\includegraphics[width=0.24\textwidth]{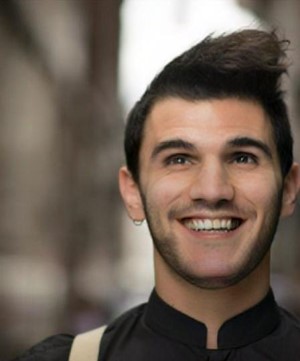}}	    
    	\subfigure[  Layer 4 + 5  ]{\includegraphics[width=0.24\textwidth]{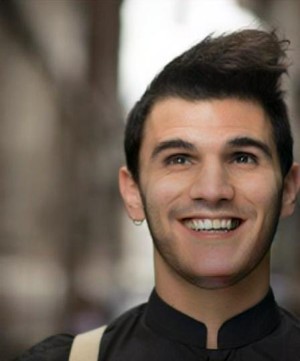}}	   
    	\subfigure[ All 3 Layers ]{\includegraphics[width=0.24\textwidth]{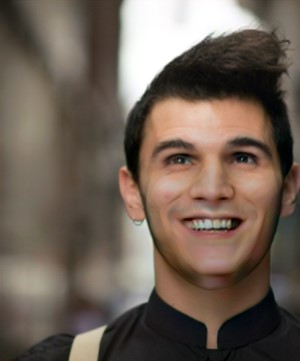}}
    	\caption{Results of removing facial hair. (a) the original image; (b), (c) and (d) are the results of using layer 5, layer 5+layer 4 and all three layers respectively. {Please zoom in to see details.}}
    	\label{fig:diff_layers}
\vspace{0.1in}
	\centering
	\subfigure[\normalsize Facial hair ]{\includegraphics[width=1\textwidth]{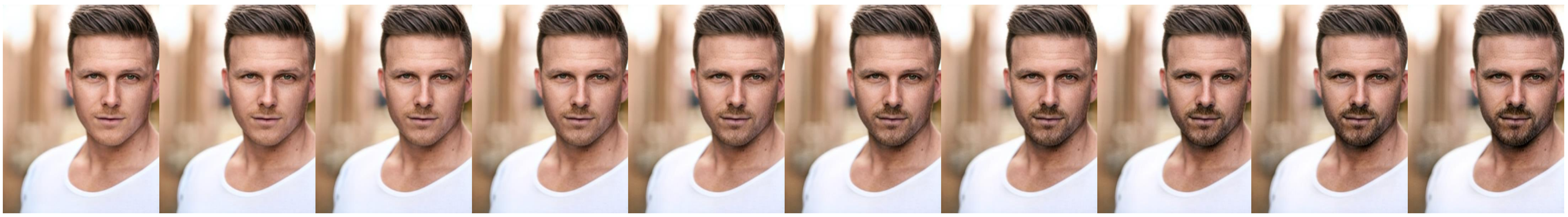}}
	\subfigure[\normalsize Smile]{\includegraphics[width=1\textwidth]{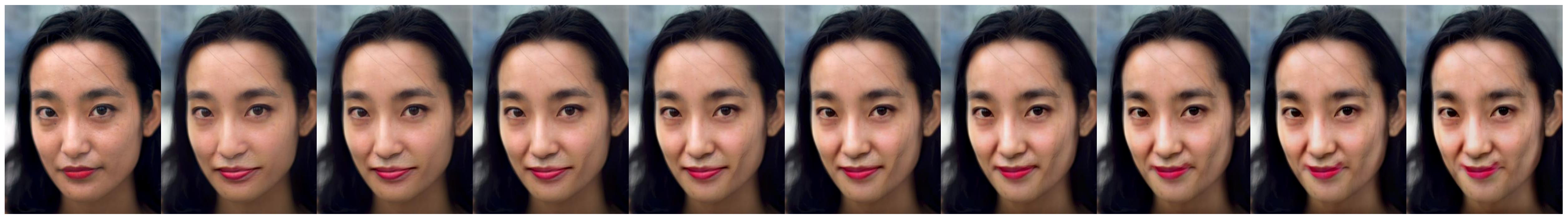}}
	\subfigure[\normalsize Younger]{\includegraphics[width=1\textwidth]{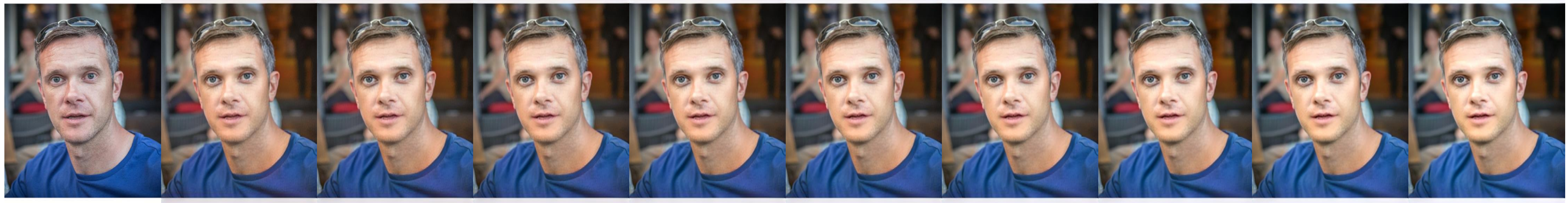}}
	\caption{Illustration of different edit strength. (a), (b) and (c) show the results of different edit strength respectively. } \label{fig:strength}
\end{figure*}
    
    \paragraph{Implicit Attention Mechanism}
    Another advantage of our Facelet Bank is the implicit attention mechanism. 
    Note that computing the pseudo target requires faces to be aligned in advance. This is not needed for our facelet bank during testing. As shown in Fig. \ref{fig:attention}, the trained operations are adaptive to the position and pose of the face, and keep staying in the correct region. This is another major advantage stemming from the fully convolution architecture. Essentially, convolution layers of our facelet bank capture the relation of certain visual patterns and corresponding $\Delta v$, which are not dependent of the position. As a result, the facelet bank is capable of detecting certain patterns (e.g., mouth for beard), and adding the effects accordingly. 
    
    \paragraph{Speed}
    Calculating $\Delta v^*$ with Eq. \eqref{eq:shift:2} is computationally expensive. The cost is mostly on face warping, deep feature extraction and nearest-neighbor search. The total testing time for a $448 \times 448$ image is 1.09 minutes with the system implemented with Pytorch, and running on a server with a Titan XP GPU and an E5-2623 v4 CPU. But for our Facelet Bank, computing $\mathcal{V}(x)$ only requires network forwarding, which costs only 0.0194 second. 
    
    \paragraph{Discussion}  
    Compared with cycleGAN \cite{CycleGAN2017}, our framework has the following advantages. First, it is very easy to train. Compared with cycleGAN that jointly optimizes two generators and two discriminators, our system only needs to learn the convolutional layers of facelet bank, which is both easier and faster. 
    
    Second, our method can handle higher-resolution images. Note that the original cycleGAN has demonstrated its effectiveness on $256 \times 256 $ images, whereas our method handles much higher resolutions, e.g., $640 \times 480$. Third, our work allows change of operation strength of effect conveniently by modifying the value of $\lambda$ in Eq. \eqref{eq:shift:2}. In contrast, changing the effect strength of cycleGAN requires the use of different training data and re-training of the model. Another related approach is DFI \cite{upchurch2016deep}, which needs no training and handles the highest resolution input data. Compared with DFI \cite{upchurch2016deep}, our approach is much faster during testing because it does not need to perform face alignment and KNN search in the deep space.

\begin{figure*}[t] 
	\centering
	\includegraphics[width=1\textwidth]{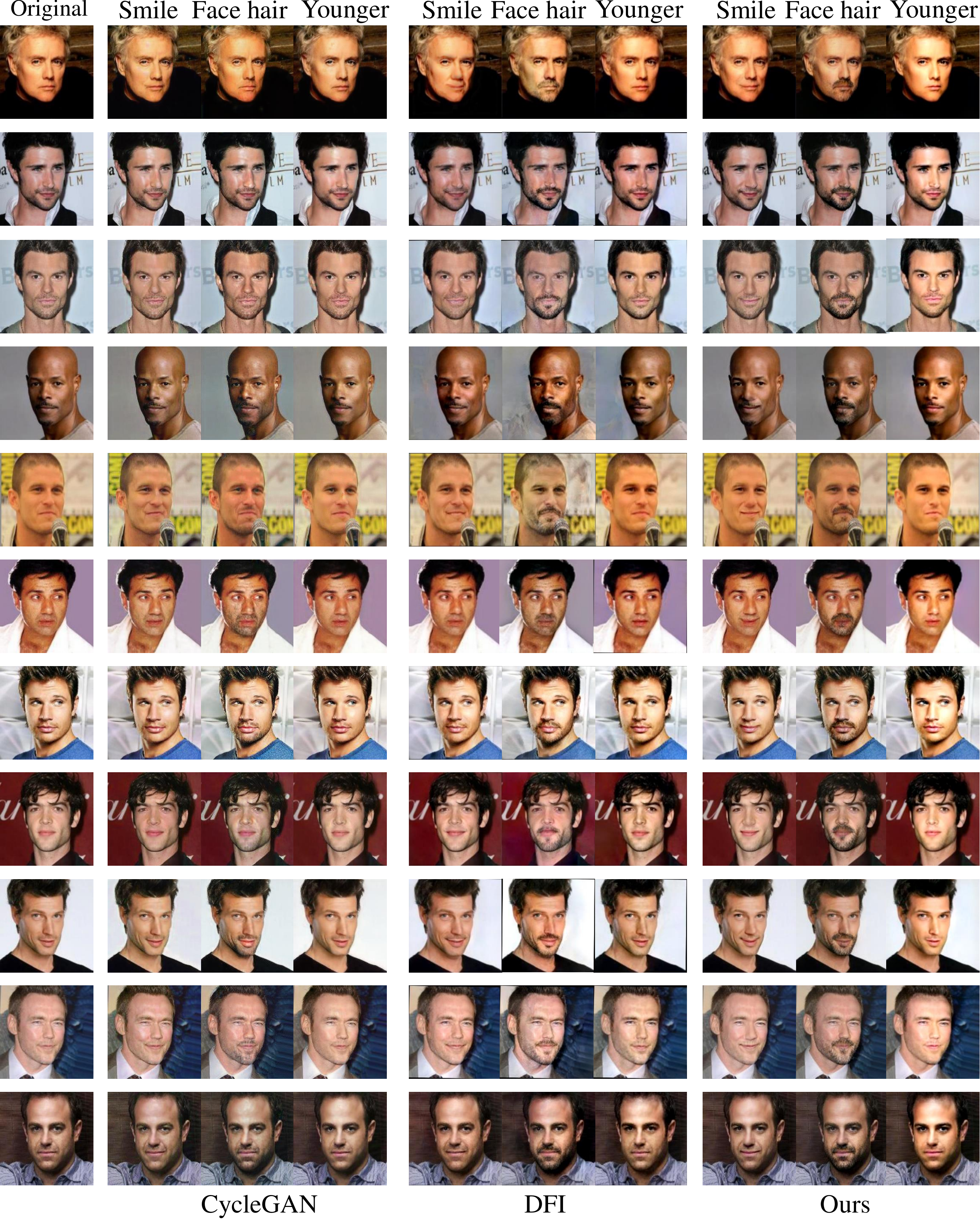}
	\caption{Comparing with CycleGAN \cite{CycleGAN2017} and DFI \cite{upchurch2016deep}. }\label{fig:compare}
\end{figure*}

    \section{Experiments}
    
    \paragraph{Implementation Details}
    We implement our model using PyTorch \cite{pytorch}. Before training the facelet bank, we train our decoder with Adam optimizer \cite{kingma2014adam} with an initial learning rate of 0.0001 and step weight decay. After training, the decoder is fixed. The face effects are trained separately. Each corresponds to a set of convolution layers of the facelet bank. Adam optimizer with default hyper-parameters are used to train the facelet-bank layers. 
    
    \paragraph{Dataset}
    Celeba \cite{liu2015faceattributes} is a large face dataset that contains 202,599 images belonging to 10,177 identities. We randomly sample $90\%$ for training and the rest are used for testing. The landmark and attribute information is used for computing the pseudo label $\Delta v^*$ computed by Eq. \eqref{eq:shift:2}. But it is not used during testing. In addition to the Celeba data, we also use the Portrait \cite{shen2016deep} and Helen \cite{le2012interactive} datasets for testing. The images of \cite{shen2016deep} and \cite{le2012interactive} are collected on Flickr with varying ages, skin colors, clothing, hair styles, etc.  We use them to validate the cross-dataset generalization ability.
    
    \subsection{Evaluation of Our Approach}
    
    The vital part of our approach is the facelet bank. In the following, we split evaluation of this design into three parts, i.e., effectiveness of the facelet bank, reason to use multi-layer aggregation, and flexibility in edit strength control.
    
    \subsubsection{Effectiveness of Facelet Bank}
    
    Estimation of attribute direction shift $\Delta v$ is critical to the result of manipulation. We compare our facelet-bank solution with several baseline approaches. The simplest way to estimate $\Delta v$ is to average both positive and negative samples, as the methods of \cite{larsen2015autoencoding, radford2015unsupervised, yan2016attribute2image, yeh2016semantic}. As shown in Fig. \ref{fig:exp1}, the beard position can be wrong since the globally computed $\Delta v$ is not adaptive to the query face. If the face pose is far from the average of the dataset, the method fails inevitably. Directly applying $\Delta v^*$ computed by Eq. \eqref{eq:delta_v_2} alleviates this problem. However, in this case, the skin color is changed since color factor is not fully canceled out by the query's nearest neighbors. Our method learns the relation between semantic visual pattern (mouth in this case) and the corresponding beard add-up operation, which cause small influence on the other regions.

    \subsubsection{Effectiveness of Multi-layer Aggregation}
    
    It was shown that different levels of layers are complementary to each other \cite{upchurch2016deep}.
    Therefore, using all of them jointly yield better results. Fig. \ref{fig:diff_layers} shows the result using different layers for the ``remove facial hair'' task. Using only layer 5 removes a large proportion of facial hair, but not complete. By incorporating layer 4 and layer 3, the amount of facial hair gradually decreases. It is eventually removed completely with all 3 layers.

    \subsubsection{Different Operation Strength}
    Our model provides a fast and convenient way to control the strength of operations. Both deep feature $\mathcal{E}(x)$ and the estimated attribute direction shift $\Delta v$ are computed only once. After that, changing the strength only requires to forward $\mathcal{E}+\lambda \Delta v$ to the decoder network. This takes only 9ms on our server. We believe that it can similarly achieve a high speed on recent high-end mobile devices. 
    
    We show results of controlling different edit strength in Fig. \ref{fig:strength}. For each case, we use strength of 0, 0.2, 0.4, 0.6, 0.8, 1, 1.2, 1.4, 1.6, and 1.8 to edit the original image. Fig. \ref{fig:strength} shows the resulting images accordingly. When $\lambda >1$, the edit can be seen as extrapolation rather than interpolation. This adds stronger edit to the original image, but sometimes yields unnatural results.
    
    \begin{figure} 
 	\centering
 	\includegraphics[width=1\linewidth]{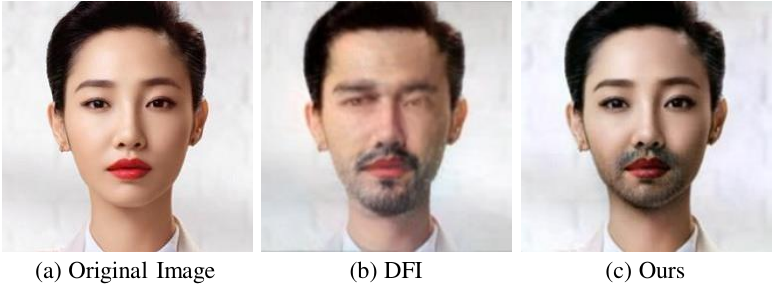}
 	\caption{Adding uncommon facial hair on a woman face;  (a) is the original image; (b) is the result of DFI \cite{upchurch2016deep}; (c) is our result. } \label{fig:lady_beard}
 \end{figure}

    \subsection{Comparison with State-of-the-Arts}
    
    Fig. \ref{fig:compare} shows the overall comparison with deep feature interpolation (DFI) \cite{upchurch2016deep} and cycleGAN \cite{CycleGAN2017}, which also perform set-to-set image attribute transform. Three effects are tested, which includes smiling, adding facial hair, and getting younger. Generally, our method can achieve better results than those of CycleGAN where the results contain stronger effect without introducing too many visual artifacts.

    Compared with the DFI, our approach achieves better performance on local effects (e.g., facial hair), since our approach introduces less irrelevant changes. For other effects, the two methods perform similarly. In terms of the amount of computation, our method is much lighter than DFI. It does not require face warping, nearest-neighbor computation and backward optimization. As a result, it is 3,371 times faster than DFI with running time of 0.0194 second and 65.4 seconds respectively. cycleGAN takes 0.0185 second for one direction, which is slightly faster than our approach. 
    
    \paragraph{Disentangling Correlated Attributes}
    We compare the results of ``adding beard to woman'' in Fig. \ref{fig:lady_beard}. Intriguingly, our model does a much better job than DFI. It is noticeable that DFI changes the overall appearance of the resulting face, while our approach maintains most of the subject's feminine characteristics. This is because DFI intrinsically is an instance-based method, which relies heavily on training samples. Since ``beard'' always comes with man's face features, it cannot disentangle the different attributes. 
    
    In contrast, our method captures the relation between ``beard'' and ``mouth''. Since the former is a masculine trait and the latter can be found in both genders, our model naturally disregards other similar relations, including the comparison of ``beard'' with ``gender''. Thus in the above example, albeit uncommon, our model yields better performance.

    \section{Concluding Remarks}
    
    In this paper, we have proposed a general framework for face manipulation. Our framework can learn face attribute shift from two image sets without any paired-example information. Thus, it does not need intensive human efforts for labeling. In addition, our framework is highly flexible -- each operation is related to only a few computed convolutional layers. 
    We have proved in our experiments that this approach yields superior results compared with other set-to-set image translation models. 
    
    {\small
    	\bibliographystyle{ieee}
    	\bibliography{egbib}
    }
    
\end{document}